\PassOptionsToPackage{numbers}{natbib}
\PassOptionsToPackage{table}{xcolor}
\documentclass{article}

\usepackage[final,main]{neurips_2026}

\usepackage[utf8]{inputenc}
\usepackage[T1]{fontenc}
\usepackage{hyperref}
\usepackage{url}
\usepackage{booktabs}
\usepackage{amsfonts}
\usepackage{amsmath}
\usepackage{amssymb}
\usepackage{nicefrac}
\usepackage{microtype}
\usepackage{xcolor}
\usepackage{graphicx}
\usepackage{subcaption}
\usepackage{multirow}
\usepackage{algorithm}
\usepackage{algorithmic}
\usepackage{cleveref}
\usepackage{xspace}
\usepackage{enumitem}
\usepackage{wrapfig}
\usepackage{placeins}

\usepackage[table]{xcolor}
\usepackage{textcomp}

\newcommand{\methodname}{\textsc{StitchEdit}\xspace}

\newcommand{\eg}{\emph{e.g.}\xspace}

\title{StitchEdit: Stitching Depth Priors into Image Editors via Per-Layer Gradient Probing}

\author{
  Qin Guo \qquad Dan Xu\thanks{Corresponding author. Project page: \url{https://guoqincode.github.io/StitchEdit/}} \\
  The Hong Kong University of Science and Technology \\
  \texttt{qguoam@connect.ust.hk} \qquad \texttt{danxu@cse.ust.hk}
}

\begin{document}

\setlength{\intextsep}{8pt plus 1pt minus 1pt}
\setlength{\textfloatsep}{12pt plus 2pt minus 2pt}
\setlength{\floatsep}{10pt plus 2pt minus 2pt}
\raggedbottom

\maketitle

\begin{abstract}
Text-guided image editing with diffusion models struggles with spatial edits (\eg manipulating objects, changing viewpoints, manipulating occlusions) that depend on the 3D structure understanding of a scene.
Marigold and follow-up work demonstrate that a pretrained diffusion backbone already harbors strong geometric priors that can be activated for depth estimation, yet how to redirect these priors toward editing remains unresolved.
Through per-layer gradient probing, we discover that the editing and depth objectives are compatible in early transformer layers but actively conflict in late layers, explaining why naive depth-conditioned editing either ignores geometry or degrades visual quality.
We present \methodname, which first activates the backbone's geometric capability via a lightweight depth-denoising adapter, then selectively fine-tunes this adapter for editing using a probe-guided per-layer learning rate schedule that preserves depth knowledge where it helps and prunes it where it conflicts.
On the proposed \emph{StitchEditBench}, a benchmark of 150 geometry-dependent edits spanning spatial transformation, occlusion manipulation, and viewpoint change, \methodname~consistently surpasses both instruction-based and depth-conditioned baselines on perceptual quality and geometric fidelity.
\end{abstract}

\begin{figure}[t]
\centering
    \includegraphics[width=0.98\textwidth]{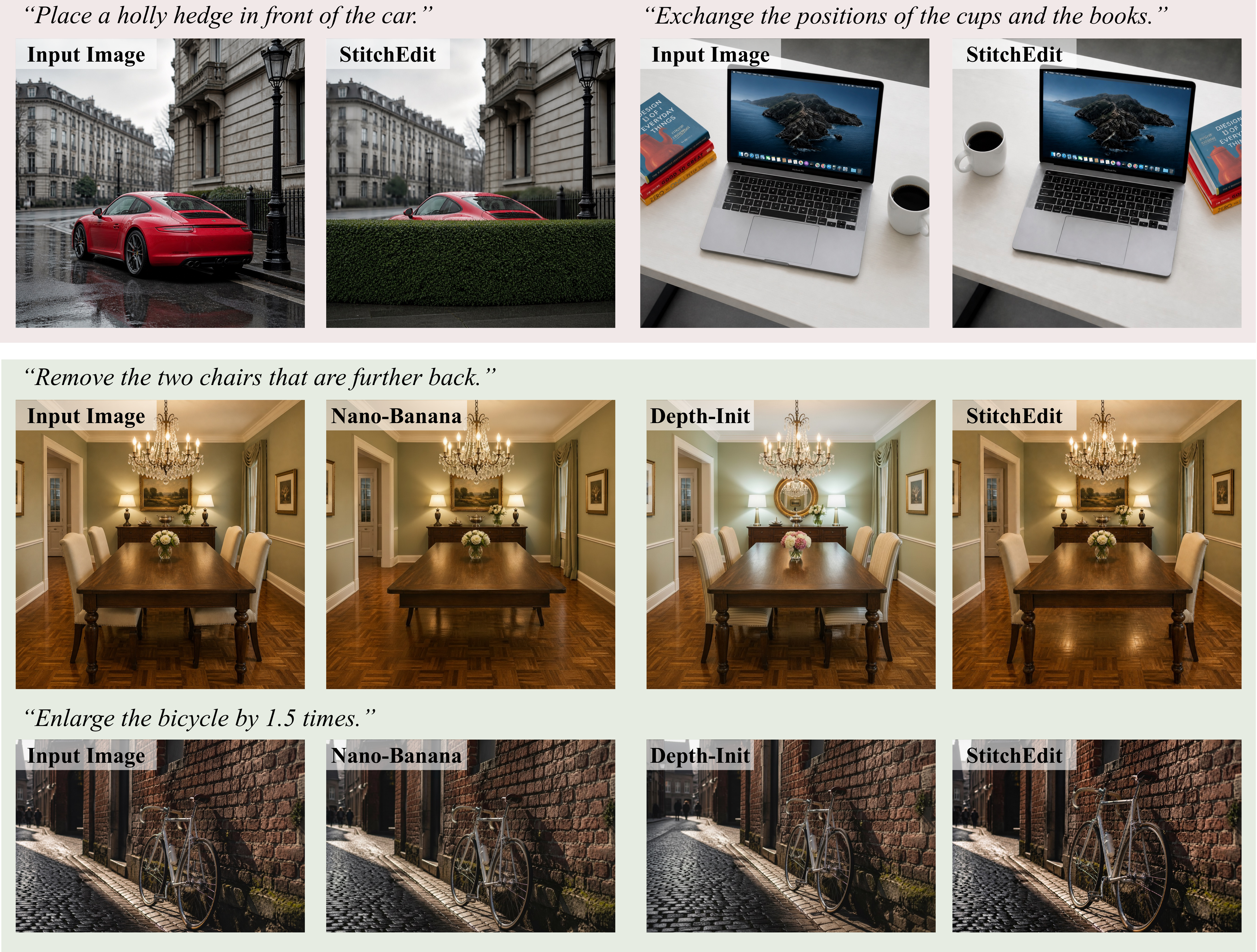}
    \caption{\textbf{Top:} Challenging 3D-aware edits generated by \methodname conditioned on text instructions. The model accurately handles complex object repositions and occlusions while strictly preserving visual quality.
    \textbf{Bottom:} Qualitative comparison against baselines. Nano-Banana~\cite{nanobanana2025} manipulates 2D appearance without explicit scene geometry, while uniformly fine-tuning the depth adapter (\textit{Depth-Init}) leads to catastrophic forgetting, causing structural failures.}
    \label{fig:teaser}
\end{figure}

\section{Introduction}
\label{sec:intro}

Moving an object closer should make it appear larger.
Inserting furniture should respect occlusion boundaries.
Changing a viewpoint should transform the entire depth field.
These spatial edits are second nature to humans yet remain difficult for text-guided editing models~\cite{brooks2023instructpix2pix,hertz2022prompt,rombach2022high,wu2025qwenimage}, which manipulate 2D appearance without explicit access to scene geometry.
The 3D knowledge required for such edits is already present inside modern diffusion models in latent form. Marigold~\cite{ke2023repurposing} and follow-up work~\cite{fu2025geowizard,he2025lotus,he2025lotus2,merge2024} show that a single diffusion backbone can be turned into a strong depth estimator with minimal adaptation, indicating that scene geometry is already represented internally as a byproduct of large-scale image generation.
We build on this premise and ask not whether to use depth, but \emph{how to transfer} the backbone's own depth capability toward the editing task.

The most natural strategy is to simply train the backbone for editing from text instructions and image-pairs.
However, without explicit or implicit geometric guidance, the resulting model falls back on 2D pattern matching and cannot enforce spatial consistency, as shown in the \textit{Nano-Banana} column of \Cref{fig:teaser}.
Another strategy is to first activate the backbone's depth capability and then continue training for spatial editing.
However, this sequential transfer also fails: the editing gradient overwrites the geometric knowledge that the depth adapter encodes, and the resulting model (\textit{Depth-Init} in \Cref{fig:teaser}) performs no better than one trained for editing from scratch.
The question, then, is \emph{which layers} can be safely fine-tuned for editing without losing the geometric features that make depth-aware editing possible.

We answer this with a per-layer gradient probe that measures, at each transformer layer, the cosine alignment between the editing gradient and the depth gradient.
This gradient probing profile is strongly positive in early transformer layers, decays through the middle, and turns negative in late layers (\Cref{fig:gradient_conflict}).
The mechanism behind the sign flip is the well-documented division of labor inside diffusion transformers~\cite{yu2025repa,wang2025ddt,xu2025ppd}: early layers carry the spatial layout that organizes a scene, where editing and depth make compatible demands, while late layers synthesize the textures and materials that render it, where the depth signal conflicts with the appearance variation editing seeks.
The implication for editing is direct: the depth adapter should be fine-tuned only where the probe is positive, and pruned entirely from the layers where it would actively degrade geometric knowledge.

We propose \methodname, a two-stage framework that turns this analysis into a concrete training strategy.
We first activate the geometric capability of the backbone with a lightweight depth-denoising adapter, confirming that the backbone already encodes rich geometric structure.
We then selectively fine-tune the resulting adapter for editing using a per-layer learning rate schedule derived from the gradient probing profile: compatible layers train at full rate, transition layers at reduced rate, and conflicting layers are pruned outright.
Because no additional gating networks or auxiliary losses are needed, the method is simple to implement and adds no overhead at inference.
On the proposed \emph{StitchEditBench}, a benchmark of 150 geometry-dependent edits spanning spatial transformation, occlusion manipulation, and viewpoint change, \methodname~consistently outperforms both instruction-based and depth-initialized baselines while preserving generation quality.

In summary, this paper makes three contributions.
\textbf{First}, we introduce per-layer gradient probing, a layer-wise measurement of cross-objective compatibility, and show that it reveals the structural cause of depth-knowledge degradation during editing fine-tuning.
\textbf{Second}, we propose \methodname, whose probe-guided training schedule selectively fine-tunes only the layers where depth and editing objectives are compatible, requiring no complex inference-time mechanisms.
\textbf{Third}, we construct \emph{StitchEditBench}, a 150-case benchmark of geometry-dependent edits, and show that \methodname~consistently outperforms instruction-based and depth-initialized baselines.

\section{Related Work}
\label{sec:related}

\textbf{Geometry-aware image editing.}
Text-guided editing methods~\cite{brooks2023instructpix2pix,hertz2022prompt,rombach2022high} operate on 2D features and lack the geometric understanding needed for spatially consistent edits.
A line of recent work injects geometry through external estimators combined with explicit 3D warping~\cite{parihar2025zero}, which introduces warping artifacts and propagates depth errors into the output, while studies of depth conditioning~\cite{geosr2024} show that models simply ignore injected geometric cues whenever conditioning does not compel their use.
\methodname~differs in two respects.
First, geometry is not injected from outside but read out of the same backbone via a lightweight adapter, so depth and editing share a representational space and require no cross-modal alignment.
Second, the resulting depth adapter is selectively fine-tuned for editing using a probe-derived per-layer schedule rather than applied uniformly, which preserves geometric knowledge where it helps and removes it where it conflicts with editing.
Related work GeoEdit~\cite{zhang2026geoedit} tackles geometric editing through 3D reconstruction and in-context inpainting with effects-sensitive attention; our method instead reads geometry out of the backbone's own representations without external 3D reconstruction.

\textbf{Layer-wise structure of diffusion transformers.}
Activating geometric capability inside a diffusion backbone has been studied for depth alone: Marigold~\cite{ke2023repurposing}, DepthFM~\cite{gui2024depthfm}, and the Lotus family~\cite{he2025lotus,he2025lotus2} repurpose generation models for depth, with parameter-efficient~\cite{hu2022lora} and unified~\cite{zhang2024jointnet,merge2024} variants preserving generation; however, all of these treat depth and generation as \emph{parallel} tasks and none use the activated depth to guide editing.
A complementary line of work characterizes the internal hierarchy of diffusion transformers: REPA~\cite{yu2025repa} shows that aligning only early transformer layers with external visual encoders outperforms uniform alignment, since unconstrained late layers retain detail-synthesis capacity, and DDT~\cite{wang2025ddt} and SP-DiT~\cite{xu2025ppd} confirm this layer-wise division of labor in decoupled and prompted DiT architectures.
\methodname~extends this hierarchy from generation and perception to geometry-aware editing: a LoRA-activated depth adapter provides geometric knowledge without error propagation, per-layer gradient probing identifies where this knowledge is compatible with editing, and probe-guided selective fine-tuning transfers depth capability to editing while preserving both geometry and generation quality.

\section{Method}
\label{sec:method}

\begin{figure}[t]
  \centering
  \includegraphics[width=1.0\linewidth]{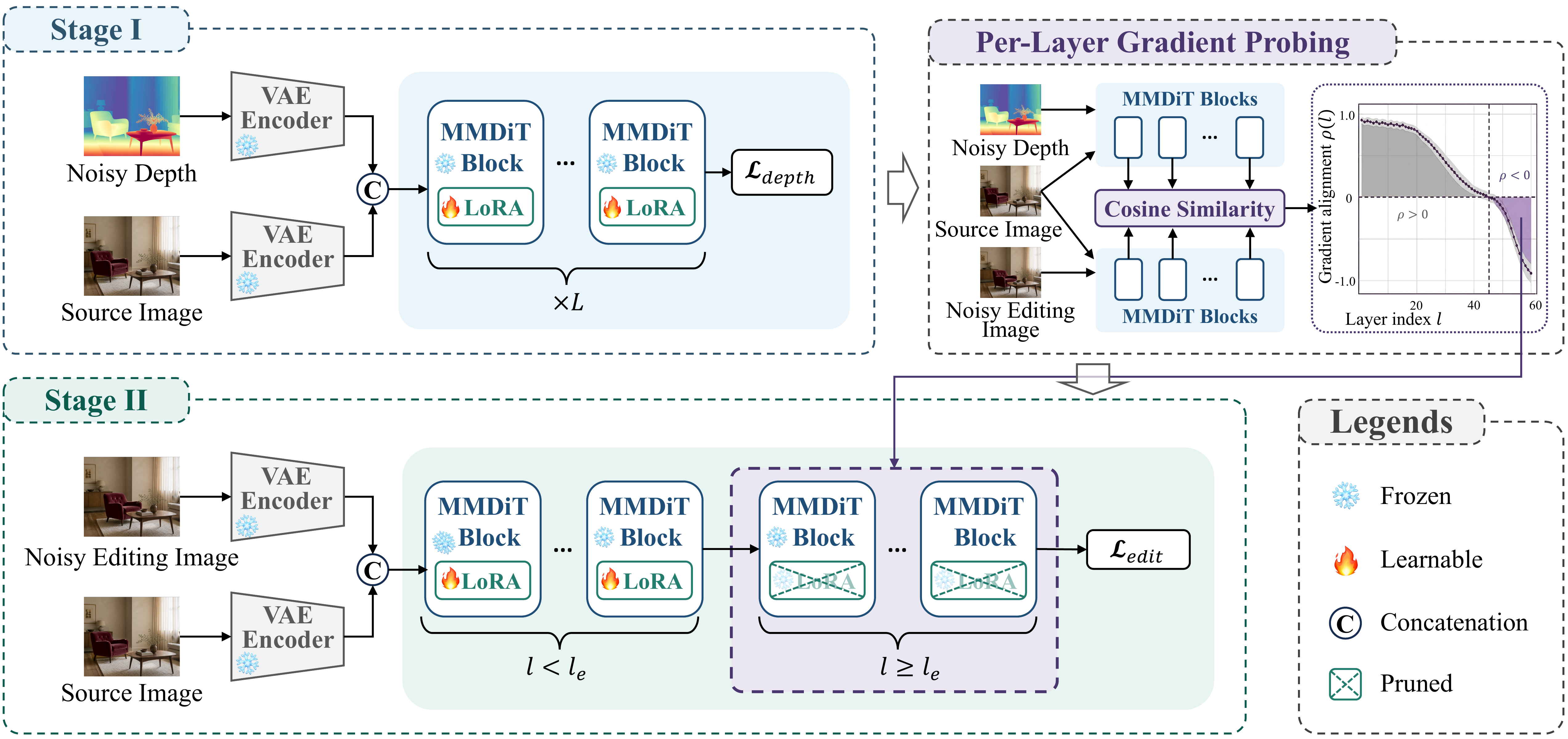}
  \caption{\textbf{Overview of \methodname.} \textbf{Stage~1}: Depth activation via a frozen-backbone LoRA adapter. \textbf{Per-layer gradient probing} reveals the layer-wise compatibility between the depth and editing objectives. \textbf{Stage~2}: The probing profile fundamentally determines both the LoRA pruning boundary and the per-layer learning rate schedule for the editing fine-tuning.}
  \label{fig:method}
\end{figure}

\begin{figure}[t]
  \centering
  \includegraphics[width=0.75\linewidth]{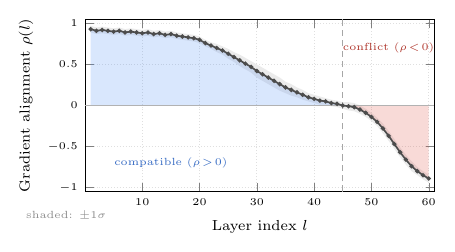}
  \caption{\textbf{Per-layer gradient probing profile $\rho(l)$.} The probe value saturates near $1$ in early layers, decays through the intermediate layers, and drops below zero in late layers, indicating that the editing gradient actively interferes with depth knowledge. The shaded band denotes $\pm$1 standard deviation across the calibration set. This empirical profile directly dictates the pruning boundary $l_{\text{e}}$ and explicitly guides the per-layer learning rate schedule in Stage~2.}
  \label{fig:gradient_conflict}
\end{figure}

\Cref{fig:method} provides an overview of \methodname.
We first introduce the preliminaries (\S\ref{sec:prelim}), then present the depth-activation stage (\S\ref{sec:probe}), the per-layer gradient probe that reveals the compatibility structure (\S\ref{sec:probe_analysis}), and ultimately the probe-guided editing fine-tuning (\S\ref{sec:guide}).

\subsection{Preliminaries}
\label{sec:prelim}

We build upon Qwen-Image-Edit~\cite{wu2025qwenimage}, a Multi-Modal Diffusion Transformer (MMDiT) backbone that maintains separate text and image branches interacting through joint attention.
For a noisy latent $\mathbf{z}_t$ at timestep $t \in [0, 1]$, the model is trained to predict the flow-matching velocity~\cite{lipman2022flow}:
\begin{equation}
\mathcal{L}_{\text{flow}} = \mathbb{E}_{t,\mathbf{z}_0,\boldsymbol{\epsilon}} \left[ \| v_{\theta}(\mathbf{z}_t, \mathbf{c}_t, \mathbf{c}, t) - (\boldsymbol{\epsilon} - \mathbf{z}_0) \|^2 \right],
\label{eq:flow}
\end{equation}
where $\mathbf{z}_t = (1-t)\mathbf{z}_0 + t\boldsymbol{\epsilon}$, $\mathbf{z}_0$ is the clean latent, and $\boldsymbol{\epsilon} \sim \mathcal{N}(\mathbf{0}, \mathbf{I})$ represents the sampled noise.
The conditioning terms $\mathbf{c}_t$ and $\mathbf{c}$ denote the text prompt and the optional reference image, respectively.
Given the velocity prediction $v_{\theta}$, the estimated clean latent can be recovered via $\hat{\mathbf{z}}_0 = \mathbf{z}_t - t \cdot v_{\theta}$.

\textbf{Functional hierarchy of transformer layers.}
A consistent empirical observation in diffusion transformers~\cite{yu2025repa,wang2025ddt,xu2025ppd} is that early layers encode the global spatial layout that structuralizes a scene, whereas later layers focus on rendering fine-grained textures and materials.
This functional division of labor provides the structural basis for the per-layer gradient probing profile we analyze in \S\ref{sec:probe_analysis}, and serves as the core motivation for our proposed layer-wise training strategy.

\subsection{Stage 1: Depth Activation via LoRA}
\label{sec:probe}

Stage~1 activates the latent depth capability of the backbone through a lightweight LoRA adapter.
We condition the backbone on the source image and train it to denoise toward the depth map.
Both the image and the depth are encoded by the shared VAE encoder ($\mathbf{z}^I=\text{Enc}(\mathbf{I})$, $\mathbf{z}^d_0=\text{Enc}(\mathbf{d})$); the noisy depth latent $\mathbf{z}^d_t$ is concatenated with the clean image latent $\mathbf{z}^I$ along the channel dimension and fed into the model.
LoRA adapters~\cite{hu2022lora} are inserted into the query, key, and value projections of each MMDiT block, keeping the backbone explicitly frozen:
\begin{equation}
\mathcal{L}_{\text{depth}} = \mathbb{E}_{t,\mathbf{z}^d_0,\boldsymbol{\epsilon}} \left[ \big\| v_{\boldsymbol{\theta}+\Delta\boldsymbol{\theta}}\big([\mathbf{z}^I,\mathbf{z}^d_t],\, t\big) - (\boldsymbol{\epsilon} - \mathbf{z}^d_0) \big\|^2 \right],
\label{eq:denoise}
\end{equation}
where $[\cdot,\cdot]$ denotes channel concatenation.
At inference, the depth latent is recovered by the same denoising process and decoded by the shared VAE decoder, requiring no additional specific decoder.
Freezing the backbone strictly preserves the generation-relevant representations upon which Stage~2 relies.

\subsection{Per-Layer Gradient Probing}
\label{sec:probe_analysis}

Following Stage~1, the LoRA adapter encodes rich geometric priors across all layers.
However, naively continuing the training process exclusively on the editing objective risks catastrophic forgetting of this knowledge, particularly in layers where the two objectives diverge.
To systematically identify these conflict zones, we introduce a per-layer gradient probe.

At each layer $l$, we quantify the compatibility between the editing and depth objectives via the cosine similarity of their respective gradients:
\begin{equation}
\rho(l) = \frac{\left\langle \nabla_{\boldsymbol{\theta}^l} \mathcal{L}_{\text{flow}}, \nabla_{\boldsymbol{\theta}^l} \mathcal{L}_{\text{depth}} \right\rangle}{\left\| \nabla_{\boldsymbol{\theta}^l} \mathcal{L}_{\text{flow}} \right\| \left\| \nabla_{\boldsymbol{\theta}^l} \mathcal{L}_{\text{depth}} \right\|},
\label{eq:probe_rho}
\end{equation}
which is computed on a calibration set of editing pairs, utilizing the frozen backbone and the depth LoRA initialized from Stage~1.
Unlike traditional hidden-state probes that extract properties from a single objective, our gradient probe explicitly measures the interference between two competing objectives at a given layer.
Using a first-order Taylor expansion, we demonstrate that updating layer $l$ with the editing gradient preserves depth knowledge only when $\rho(l) > 0$. Conversely, when $\rho(l) < 0$, training for editing actively increases the depth loss, thereby degrading geometric features:
\begin{equation}
\Delta \mathcal{L}_{\text{depth}} \approx -\eta \left\| \nabla_{\boldsymbol{\theta}^l} \mathcal{L}_{\text{flow}} \right\| \left\| \nabla_{\boldsymbol{\theta}^l} \mathcal{L}_{\text{depth}} \right\| \rho(l),
\label{eq:conflict}
\end{equation}
where $\eta$ denotes the learning rate.

Empirically (\Cref{fig:gradient_conflict}), we observe that $\rho(l)$ saturates near $1$ in the early MMDiT layers, gradually decays through the intermediate layers, and ultimately crosses zero in the late layers. The shaded confidence band narrows at both extremes, confirming that alignment and conflict are consistent properties of their respective layer groups, while the transition zone exhibits greater variance across calibration pairs. This trajectory reveals three distinct regimes:
\textbf{(1) Compatible zone} ($l \leq l_{\text{s}}$): $\rho \approx 1$; objectives reinforce each other.
\textbf{(2) Transition zone} ($l_{\text{s}} < l < l_{\text{e}}$): $\rho$ decays toward zero; objectives become increasingly orthogonal.
\textbf{(3) Conflict zone} ($l \geq l_{\text{e}}$): $\rho < 0$; editing actively degrades depth knowledge.
This gradient profile provides a principled basis for determining both the pruning boundary and the per-layer learning rate schedule in Stage~2.

\subsection{Stage 2: Probe-Guided Editing Fine-tuning}
\label{sec:guide}

We continue training the Stage~1 LoRA for the editing task, modulating the per-layer training intensity according to the probing profile. This ensures that geometric priors are preserved where they are synergistic with editing, and rigorously pruned where they induce conflict.

\textbf{LoRA pruning.}
We remove the LoRA adapters from all layers $l \geq l_{\text{e}}$, where $\rho(l) < 0$.
These layers revert to the frozen backbone behavior, ensuring that the appearance-synthesizing capacity of the late layers is not corrupted by depth-trained weights whose gradients explicitly oppose the editing objective.

\textbf{Per-layer learning rate schedule.}
For the retained layers ($l < l_{\text{e}}$), we modulate the learning rate using a probe-derived cosine multiplier:
\begin{equation}
\eta(l) = w(l) \cdot \eta_0, \quad
w(l) = \mathbb{1}[l \leq l_{\text{s}}] + \cos\!\left(\tfrac{\pi}{2} \cdot \tfrac{l - l_{\text{s}}}{l_{\text{e}} - l_{\text{s}}}\right) \mathbb{1}[l_{\text{s}} < l < l_{\text{e}}],
\label{eq:schedule}
\end{equation}
such that layers with high $\rho$ train at the full rate, while transition layers receive progressively attenuated rates. The boundary $l_{\text{s}}$ marks the onset of decay in the gradient probing profile.

\textbf{Training objective.}
The editing model takes as input the clean source-image latent $\mathbf{z}^I$ and the noisy target-edit latent $\mathbf{z}^e_t$, concatenated along the channel dimension, together with the text instruction $\mathbf{c}_t$:
\begin{equation}
\mathcal{L_{\text{depth}}} = \mathbb{E}_{t,\mathbf{z}^e_0,\boldsymbol{\epsilon}} \left[ \big\| v_{\boldsymbol{\theta}+\Delta\boldsymbol{\theta}'}([\mathbf{z}^I,\mathbf{z}^e_t],\, \mathbf{c}_t,\, t) - (\boldsymbol{\epsilon} - \mathbf{z}^e_0) \big\|^2 \right],
\label{eq:edit_loss}
\end{equation}
where $\Delta\boldsymbol{\theta}'$ denotes the LoRA parameters after pruning.
Our design is intentionally minimal: there is no dual forward pass, no per-layer gating network, and no auxiliary geometric loss. The probe-guided schedule alone optimally balances geometry and editing.

\subsection{StitchEditBench: Task Taxonomy and Data Construction}
\label{sec:benchmark}

We introduce \textbf{StitchEditBench}, a comprehensive benchmark organized around geometry-aware edits. Each sample is formulated as a 5-tuple: $(\mathbf{I}_{\text{src}}, \mathbf{c}_t, \mathbf{I}_{\text{tgt}}, \mathbf{D}_{\text{src}}, \mathbf{D}_{\text{tgt}})$, representing the source image, text instruction, target image, source depth, and target depth, respectively.

\textbf{Task taxonomy.}
We categorize geometric edits into three types: \textbf{(1) Spatial Transformation (ST)} repositions objects in 3D space (translate, rotate, rescale), requiring adjustments for scale and shadows; \textbf{(2) Occlusion Manipulation (OM)} alters visibility relationships (insert, remove, reorder), requiring rigorous depth ordering and occluded-region completion; and \textbf{(3) Viewpoint Change (VC)} applies camera transformations, requiring global structural transformation of the depth field.

\textbf{Data construction \& Statistics.}
We synthesize paired data by chaining a Large Language Model planner (Gemini~\cite{gemini2025}) and an image generator (Nano-Banana~\cite{nanobanana2025}). Specifically, Gemini is prompted with a source image and a category label to produce a structured edit specification alongside a natural-language instruction $\mathbf{c}_t$. The same Gemini model acts as a secondary verifier to filter out non-compliant geometric transformations, yielding a high-precision pool. The depth maps ($\mathbf{D}_{\text{src}}, \mathbf{D}_{\text{tgt}}$) are generated by our Stage~1 depth model.
The training set comprises 10K pairs (4K ST, 3.5K OM, 2.5K VC). The evaluation benchmark contains 150 meticulously curated cases drawn from a strictly held-out, non-overlapping source-image pool.

\section{Experiments}
\label{sec:experiments}

\subsection{Experimental Setup}
\label{sec:experimental_setup}

\textbf{Implementation Details.}
Our framework is built upon Qwen-Image-Edit~\cite{wu2025qwenimage}, a DiT backbone comprising $L\!=\!60$ blocks.
We attach LoRA adapters (rank $r\!=\!64$, $\alpha\!=\!128$) to the query, key, and value projections of each MMDiT block.
All images are processed at $1024\!\times\!1024$ resolution with classifier-free guidance scale 4.0. All experiments are conducted on 8 NVIDIA H100 GPUs.
\textbf{Stage~1} trains the depth LoRA for 50K steps at a batch size of 32 using AdamW (learning rate $1\!\times\!10^{-4}$, weight decay $1\!\times\!10^{-2}$, 1K-step warmup) on the 3M image-depth pairs curated by Lotus-2~\cite{he2025lotus2}.
\textbf{Per-layer gradient probing} is computed on a 1K-pair calibration subset drawn from the Stage~2 training pool, yielding the critical boundaries $l_{\text{s}}\!=\!20$ and $l_{\text{e}}\!=\!45$.
\textbf{Stage~2} consequently prunes the LoRA from layers $l \geq 45$ and continues fine-tuning on our constructed 10K editing pairs. This stage trains for 100K steps at a batch size of 16 with AdamW (learning rate $2\!\times\!10^{-5}$, weight decay $1\!\times\!10^{-2}$, 500-step warmup), strictly applying the probe-guided schedule formulated in \Cref{eq:schedule}.

\begin{figure}[t]
  \centering
  \includegraphics[width=1.0\linewidth]{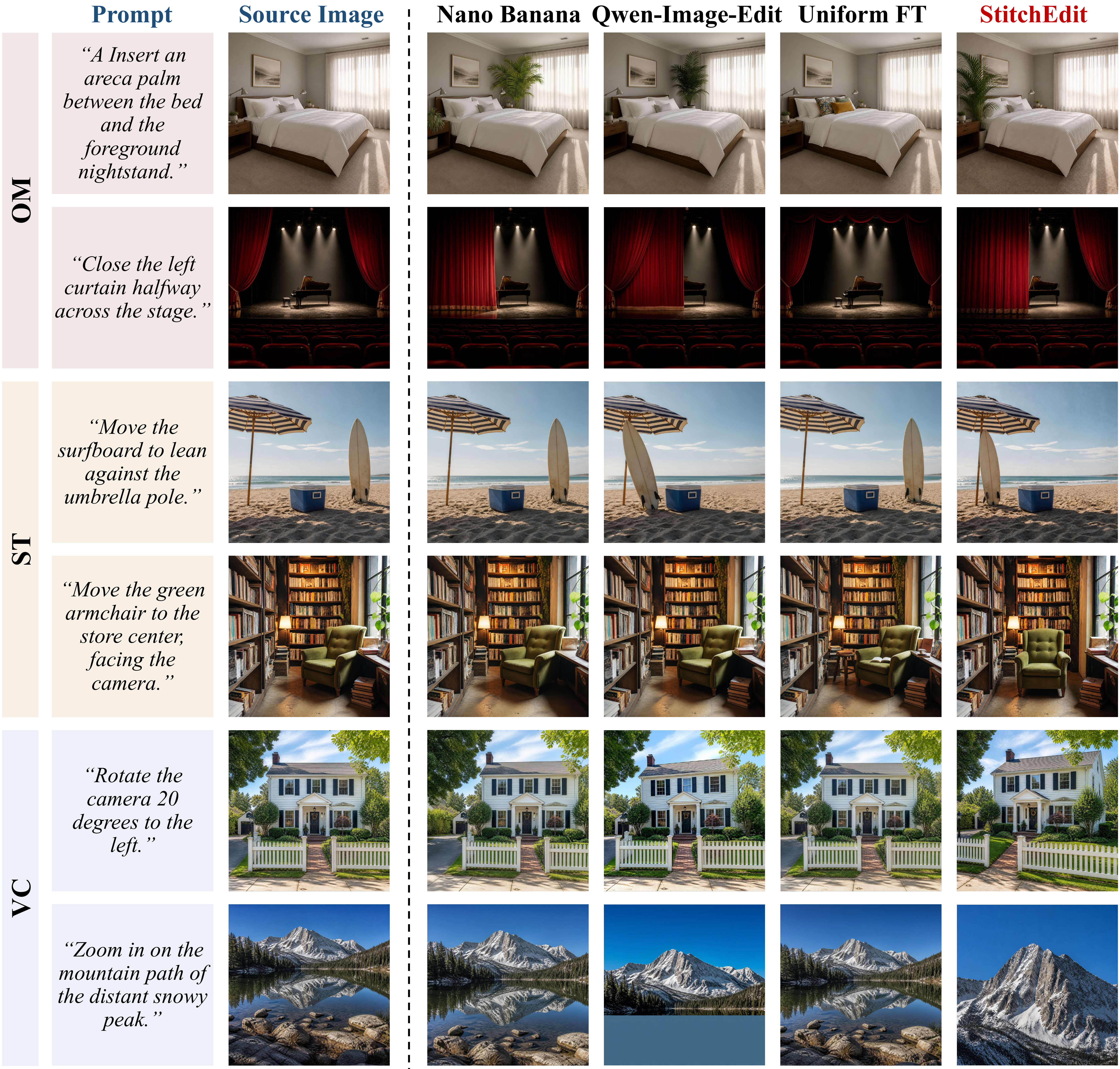}
  \caption{\textbf{Qualitative comparison.} \methodname~maintains strict geometric consistency while preserving exceptional visual quality. Instruction-based editors frequently fail to respect 3D spatial structures, and Uniform FT suffers from depth degradation.}
  \label{fig:qualitative}
\end{figure}

\textbf{Benchmark and Metrics.}
We evaluate our method on the proposed \emph{StitchEditBench}. 
Perceptual quality is measured by \textit{LPIPS}~\cite{zhang2018perceptual}.
Geometric fidelity is evaluated via the \textit{Depth Consistency Score (DCS)}, defined as the Pearson correlation between the target depth and the depth of the edited output (estimated by Depth Anything V2~\cite{depth_anything_v2}).
Instruction following is measured by \textit{CLIP-Dir}~\cite{radford2021learning}, which calculates the cosine similarity between the CLIP embedding change of the image pair and the instruction embedding.
Additionally, we report a \textit{User Preference} score, aggregating rankings from 50 human participants across all test cases.

\textbf{Baselines.}
We benchmark against two distinct families of methods.
\textit{Instruction-based editors} rely solely on the source image and textual instructions, represented by the state-of-the-art closed-source model Nano-Banana~\cite{nanobanana2025} and the open-source model Qwen-Image-Edit~\cite{wu2025qwenimage}.
\textit{Depth-initialized editors} benefit from our Stage~1 depth pre-training. We define \textit{Uniform FT} as a rigorous ablation baseline where all LoRA layers are trained uniformly at a constant learning rate, devoid of our probe guidance.

\begin{table}[!htbp]
\centering
\caption{\textbf{Main results on \emph{StitchEditBench}.} Uniform FT starts from the same depth-pre-trained adapter as \methodname~but trains all layers uniformly, leading to the loss of geometric knowledge. \methodname~resolves this via probe-guided selective training.}
\label{tab:main}
\small
\setlength{\tabcolsep}{6pt}
\begin{tabular}{lcccc}
\toprule
Method & LPIPS$\downarrow$ & DCS$\uparrow$ & CLIP-Dir$\uparrow$ & User Pref.$\uparrow$ \\
\midrule
\multicolumn{5}{l}{\textit{Instruction-based}} \\[2pt]
Nano-Banana~\cite{nanobanana2025} & 0.13 & 0.74 & 0.78 & 16\% \\
Qwen-Image-Edit~\cite{wu2025qwenimage} & 0.14 & 0.71 & 0.76 & 14\% \\
\midrule
\multicolumn{5}{l}{\textit{Depth-initialized}} \\[2pt]
Uniform FT & 0.14 & 0.80 & 0.76 & 12\% \\
\midrule
\rowcolor{gray!10}
\textbf{\methodname~(Ours)} & \textbf{0.10} & \textbf{0.92} & \textbf{0.81} & \textbf{58\%} \\
\bottomrule
\end{tabular}
\end{table}

\subsection{Main Results}
\label{sec:main_results}

\Cref{tab:main} reports the quantitative comparison across all 150 test cases.
\methodname~consistently achieves the best performance across all metrics.
Notably, while Nano-Banana and Qwen-Image-Edit yield visually plausible outputs, they score poorly on DCS, confirming that semantic text supervision alone is insufficient to enforce strict geometric consistency.
Uniform FT serves as the most revealing depth-initialized baseline: despite inheriting the same depth-pre-trained adapter, uniformly updating all layers leads to catastrophic forgetting of geometric knowledge in conflict zones, degrading the DCS to 0.80.
Conversely, the probe-guided schedule of \methodname~optimally preserves geometric priors where they cooperate with editing and prunes them where they interfere, unlocking the best trade-off between visual quality and geometric fidelity.
The 58\% User Preference score further corroborates this: human raters consistently favor outputs that respect spatial plausibility, and the 0.12-point DCS gap between Uniform FT and \methodname~(0.80 vs.\ 0.92) confirms that the late-layer conflict zone identified by our probe is the primary source of geometric forgetting.

\textbf{Per-category analysis.}
The superiority of \methodname~scales proportionally with the geometric complexity of the edit: relative to the strongest instruction-based baseline, our method reduces LPIPS by $0.03$ on Spatial Transformation (ST), $0.04$ on Viewpoint Change (VC), and $0.02$ on Occlusion Manipulation (OM). The DCS performance gap remains consistently substantial across all categories (\Cref{tab:category}).
The largest DCS improvement appears on VC (0.90 vs.\ 0.69 for Nano-Banana), where global depth-field transformation demands coherent geometric reasoning across the entire image.
\Cref{fig:qualitative} provides representative qualitative comparisons: instruction-based editors frequently disregard scale transformations and occlusion logic; Uniform FT partially corrupts depth constraints; whereas \methodname~synthesizes geometrically rigorous and visually compelling edits.

\begin{table}[!htbp]
\centering
\caption{\textbf{Per-category breakdown on StitchEditBench.} All three categories require geometric understanding; gains are consistent across spatial transformation, occlusion manipulation, and viewpoint change.}
\label{tab:category}
\small
\setlength{\tabcolsep}{5pt}
\begin{tabular}{l cc cc cc}
\toprule
& \multicolumn{2}{c}{ST (60)} & \multicolumn{2}{c}{OM (50)} & \multicolumn{2}{c}{VC (40)} \\
\cmidrule(lr){2-3} \cmidrule(lr){4-5} \cmidrule(lr){6-7}
Method & LPIPS$\downarrow$ & DCS$\uparrow$ & LPIPS$\downarrow$ & DCS$\uparrow$ & LPIPS$\downarrow$ & DCS$\uparrow$ \\
\midrule
Nano-Banana~\cite{nanobanana2025} & 0.14 & 0.71 & 0.13 & 0.76 & 0.16 & 0.69 \\
Qwen-Image-Edit~\cite{wu2025qwenimage} & 0.15 & 0.68 & 0.14 & 0.73 & 0.17 & 0.66 \\
Uniform FT & 0.15 & 0.78 & 0.14 & 0.81 & 0.16 & 0.77 \\
\midrule
\rowcolor{gray!10}
\textbf{\methodname} & \textbf{0.11} & \textbf{0.91} & \textbf{0.11} & \textbf{0.91} & \textbf{0.12} & \textbf{0.90} \\
\bottomrule
\end{tabular}
\end{table}

\subsection{Ablation Studies}
\label{sec:ablation}

\textbf{Training strategy.}
\Cref{tab:schedule} validates the efficacy of our probe-guided schedule against alternative strategies.
Direct LoRA fine-tuning without Stage~1 depth pre-training yields the poorest DCS (0.70), confirming that explicit geometric initialization is indispensable.
Uniform FT improves DCS to 0.80 but still suffers from geometric forgetting in the late layers.
Top-$K$ Freeze retains depth LoRA only in the first $K$ layers and removes it elsewhere; its 0.87 DCS shows that a hard cutoff discards useful geometric information in the transition zone.
Random Schedule assigns each layer a uniformly sampled learning rate multiplier, serving as a controlled baseline to verify that the probe's layer ordering carries genuine signal rather than benefiting from mere variance in training intensity; its 0.83 DCS confirms this.
A heuristic Fixed Cosine decay captures a significant portion of the benefit (0.90 DCS), validating the existence of the three-regime structure, but its fixed boundaries cannot adapt to the actual probe profile.
Ultimately, our dynamically computed \textbf{Probe-Guided schedule} achieves the apex performance (0.92 DCS, 0.10 LPIPS) by precisely calibrating the learning intensity to the layer-wise objective compatibility.

\begin{table}[H]
\centering
\caption{\textbf{Ablation on training strategy.} Probe-guided selective fine-tuning significantly outperforms both uniform training and heuristic schedules. Direct fine-tuning validates the absolute necessity of Stage~1 geometric initialization.}
\label{tab:schedule}
\small
\setlength{\tabcolsep}{6pt}
\begin{tabular}{lcccc}
\toprule
Strategy & Depth Init & LPIPS$\downarrow$ & DCS$\uparrow$ & CLIP-Dir$\uparrow$ \\
\midrule
Direct LoRA & \texttimes & 0.17 & 0.70 & 0.74 \\
\midrule
Uniform FT & \checkmark & 0.14 & 0.80 & 0.76 \\
Top-$K$ Freeze & \checkmark & 0.12 & 0.87 & 0.78 \\
Fixed Cosine & \checkmark & 0.11 & 0.90 & 0.80 \\
Random Schedule & \checkmark & 0.13 & 0.83 & 0.77 \\
\rowcolor{gray!10}
\textbf{Probe-Guided (Ours)} & \checkmark & \textbf{0.10} & \textbf{0.92} & \textbf{0.81} \\
\bottomrule
\end{tabular}
\end{table}

\begin{wraptable}{r}{0.46\linewidth}
\vspace{-\intextsep}
\captionsetup{skip=4pt}
\centering
\caption{\textbf{Explicit vs.\ implicit conditioning.}}
\label{tab:explicit}
\small
\setlength{\tabcolsep}{3pt}
\begin{tabular}{lccc}
\toprule
Method & LPIPS$\downarrow$ & DCS$\uparrow$ & CLIP-Dir$\uparrow$ \\
\midrule
CN-D ($\mathbf{D}_{\text{src}}$) & 0.17 & 0.72 & 0.62 \\
CN-D ($\mathbf{D}_{\text{tgt}}$)$^\dagger$ & 0.12 & 0.90 & 0.79 \\
\midrule
\rowcolor{gray!10}
\textbf{Ours} & \textbf{0.10} & \textbf{0.92} & \textbf{0.81} \\
\bottomrule
\multicolumn{4}{l}{\scriptsize $^\dagger$Oracle: target depth unavailable.}
\end{tabular}
\end{wraptable}
\textbf{Explicit vs.\ implicit geometric conditioning.}
A natural alternative to our implicit depth-prior approach is to supply an explicit depth map through a ControlNet-style adapter~\cite{zhang2023adding}.
We evaluate two variants: \textit{ControlNet\nobreakdash-Depth ($\mathbf{D}_{\text{src}}$)} conditions the backbone on the \emph{source} depth map, and \textit{ControlNet\nobreakdash-Depth ($\mathbf{D}_{\text{tgt}}$)} conditions on the \emph{target} depth map.
The latter serves as an \textbf{oracle upper bound}, as the target depth is strictly unavailable at inference time.

\Cref{tab:explicit} reveals the core limitation of explicit conditioning.
Conditioning on the source depth rigidly locks the spatial layout to the initial geometry, causing the model to resist the structural changes demanded by the edit instruction. This leads to a severe degradation in both instruction following (0.62 CLIP\nobreakdash-Dir) and geometric fidelity (0.72 DCS, measured against the \emph{target} depth).
Even the oracle variant, which receives the ground-truth target depth, achieves only 0.90 DCS and 0.79 CLIP\nobreakdash-Dir, still underperforming \methodname~(0.92, 0.81).
This performance gap arises because rigid ControlNet conditioning excessively constrains the network's appearance-synthesis capacity, whereas \methodname's implicit prior leaves the backbone free to jointly optimize geometry and visual quality.
In particular, the oracle variant demonstrates that even perfect geometric input cannot compensate for the loss of representational flexibility: the ControlNet signal competes with the backbone's own features at every layer, whereas our probe-guided LoRA cooperates with the backbone in early layers and simply stays silent in late ones.

\begin{wraptable}{r}{0.46\linewidth}
\vspace{-\intextsep}
\captionsetup{skip=4pt}
\centering
\caption{\textbf{Sensitivity to $l_{\text{s}},l_{\text{e}}$ boundaries.}}
\label{tab:boundary}
\small
\setlength{\tabcolsep}{3pt}
\begin{tabular}{cc ccc}
\toprule
$l_{\text{s}}$ & $l_{\text{e}}$ & LPIPS$\downarrow$ & DCS$\uparrow$ & CLIP-Dir$\uparrow$ \\
\midrule
10 & 35 & 0.12 & 0.86 & 0.78 \\
15 & 40 & 0.11 & 0.89 & 0.79 \\
\rowcolor{gray!10}
\textbf{20} & \textbf{45} & \textbf{0.10} & \textbf{0.92} & \textbf{0.81} \\
25 & 50 & 0.11 & 0.91 & 0.78 \\
30 & 55 & 0.13 & 0.92 & 0.75 \\
\bottomrule
\end{tabular}
\end{wraptable}
\textbf{Schedule boundary sensitivity.}
\Cref{tab:boundary} varies $l_{\text{s}}$ and $l_{\text{e}}$ around the probed defaults.
Performance peaks at $(20, 45)$ and degrades when the compatible region is too narrow ($l_{\text{s}}=10$: insufficient geometric capacity) or too broad ($l_{\text{s}}=30$: conflicting layers are trained, degrading depth knowledge).
The optimal $l_{\text{s}}=20$ aligns with the REPA~\cite{yu2025repa} finding that roughly the first third of layers carry spatial structure, suggesting a shared structural boundary in diffusion transformers.
Notably, the $(30, 55)$ setting maintains high DCS (0.92) but suffers a sharp CLIP-Dir drop to 0.75, indicating that training conflict-zone layers preserves depth at the cost of editing fidelity---precisely the trade-off our probe is designed to avoid.

\begin{wraptable}{r}{0.46\linewidth}
\vspace{-\intextsep}
\captionsetup{skip=4pt}
\centering
\caption{\textbf{Decay function shape.}}
\label{tab:decay}
\small
\setlength{\tabcolsep}{4pt}
\begin{tabular}{lccc}
\toprule
Decay & LPIPS$\downarrow$ & DCS$\uparrow$ & CLIP-Dir$\uparrow$ \\
\midrule
Step & 0.13 & 0.88 & 0.78 \\
Linear & 0.11 & 0.90 & 0.79 \\
\rowcolor{gray!10}
\textbf{Cosine} & \textbf{0.10} & \textbf{0.92} & \textbf{0.81} \\
\bottomrule
\end{tabular}
\end{wraptable}
\textbf{Decay function.}
We evaluate three mathematical decay functions for the transition zone ($l_{\text{s}} < l < l_{\text{e}}$).
As shown in \Cref{tab:decay}, the Cosine decay formulation achieves the best balance (0.10 LPIPS, 0.92 DCS), outperforming both step-function (0.13, 0.88) and linear decay (0.11, 0.90). Its slow-fast-slow continuous profile strategically concentrates parameter updates at the stable regime extremes while rapidly traversing the ambiguous middle layers.

\begin{wraptable}{r}{0.46\linewidth}
\vspace{-\intextsep}
\captionsetup{skip=4pt}
\centering
\caption{\textbf{Depth estimation (NYU V2).}}
\label{tab:depth}
\small
\setlength{\tabcolsep}{4pt}
\begin{tabular}{lccc}
\toprule
Method & AbsRel$\downarrow$ & $\delta_1$$\uparrow$ & Edit \\
\midrule
Marigold & 5.5 & 0.964 & \texttimes \\
Depth Any. V2 & 4.3 & 0.981 & \texttimes \\
Qwen-IE & 12.2 & 0.865 & \checkmark \\
\rowcolor{gray!10}
\textbf{Ours (S1)} & \textbf{5.4} & \textbf{0.966} & \checkmark \\
\bottomrule
\end{tabular}
\end{wraptable}
\textbf{Depth estimation capacity.}
Beyond its primary role in facilitating editing, the Stage~1 depth LoRA alone achieves competitive depth estimation performance (\Cref{tab:depth}). On the NYU Depth V2 dataset~\cite{silberman2012indoor}, our model attains 5.4 AbsRel, slightly outperforming dedicated depth estimators (\eg Marigold~\cite{ke2023repurposing}: 5.5). Notably, Qwen-Image-Edit's 12.2 AbsRel reveals that the backbone's latent geometric knowledge is poorly activated without dedicated depth training; the Stage~1 LoRA, despite its lightweight parameter footprint, reduces the error by more than half (5.4 vs.\ 12.2), confirming that a small adapter suffices to unlock the backbone's depth capability.

\section{Conclusion and Limitations}
\label{sec:conclusion}

We presented \methodname to solve the challenge of spatial manipulations in text-guided image editing. Through per-layer gradient probing, we found that depth and editing objectives cooperate in early transformer layers but actively conflict in late ones. This insight drives a selective fine-tuning schedule that preserves geometric knowledge where it helps and prunes it where it harms. As a result, \methodname achieves strict spatial consistency without the rigidity of explicit depth conditioning.

\textbf{Limitations.}
\methodname~currently relies on natural-language instructions to specify spatial edits, which limits the precision of geometric control. For instance, the user cannot prescribe an exact scale factor or rotation angle; the model interprets approximate intent from text. Extending the framework with explicit geometric parameters (\eg rotation angles, translation offsets) to enable fine-grained spatial control is a promising direction for future work.

\FloatBarrier
\newpage

\section*{Acknowledgements}
The work is supported in part by the Early Career Scheme of the Research Grants Council (RGC) of the Hong Kong SAR under grant No. 26202321, ITF PRP/046/24FX, the Science \& Technology Cooperation Program of Shandong under grant No. SDST26EG01, SAIL Research Project, and HKUST Zeekr Collaborative Research Fund.

\bibliographystyle{plain}
\bibliography{references}

\par\bigskip

\appendix

\section{Per-Layer Gradient Probing Details}

The probing profile $\rho(l)$ is computed on a 1K-pair calibration subset drawn from the Stage~2 training pool.
For each pair, we compute $\nabla_{\boldsymbol{\theta}^l}\mathcal{L}_{\text{flow}}$ and $\nabla_{\boldsymbol{\theta}^l}\mathcal{L}_{\text{depth}}$ at every MMDiT block using the frozen backbone and the Stage~1 LoRA, then take the per-layer cosine similarity and average across the calibration set.
Probing takes ${\sim}$30 minutes on a single H100 and is run once before Stage~2 training; the resulting curve (\Cref{fig:gradient_conflict}) determines $l_{\text{s}}$ and $l_{\text{e}}$.

\section{Broader Impact Statement}

\methodname~enables geometrically plausible editing for creative applications, VR/AR, and content generation.
As with all editing methods, potential misuse for generating misleading content exists.
We encourage responsible use and development of detection methods for edited images.

\newpage
\section*{NeurIPS Paper Checklist}

\begin{enumerate}

\item {\bf Claims}
    \item[] Question: Do the main claims made in the abstract and introduction accurately reflect the paper's contributions and scope?
    \item[] Answer: \answerYes{}
    \item[] Justification: The abstract and introduction state three contributions (per-layer gradient probing, probe-guided selective fine-tuning, StitchEditBench), all of which are supported by the experiments in \S\ref{sec:experiments} and appendix.

\item {\bf Limitations}
    \item[] Question: Does the paper discuss the limitations of the work performed by the authors?
    \item[] Answer: \answerYes{}
    \item[] Justification: Limitations are discussed.

\item {\bf Theory assumptions and proofs}
    \item[] Question: For each theoretical result, does the paper provide the full set of assumptions and a complete (and correct) proof?
    \item[] Answer: \answerNA{}
    \item[] Justification: The paper does not present formal theorems. The first-order gradient analysis in \Cref{eq:conflict} is a standard Taylor expansion used as motivation, not a formal theorem.

    \item {\bf Experimental result reproducibility}
    \item[] Question: Does the paper fully disclose all the information needed to reproduce the main experimental results of the paper to the extent that it affects the main claims and/or conclusions of the paper (regardless of whether the code and data are provided or not)?
    \item[] Answer: \answerYes{}
    \item[] Justification: All hyperparameters, training schedules, data construction pipelines, and evaluation metrics are detailed in \S\ref{sec:experiments} and the appendix. The probing boundary values ($l_s$, $l_e$) and the per-layer learning rate schedule are fully specified.

\item {\bf Open access to data and code}
    \item[] Question: Does the paper provide open access to the data and code, with sufficient instructions to faithfully reproduce the main experimental results, as described in supplemental material?
    \item[] Answer: \answerNo{}
    \item[] Justification: Code and the StitchEditBench dataset will be released upon acceptance. The paper provides sufficient detail for independent reimplementation.

\item {\bf Experimental setting/details}
    \item[] Question: Does the paper specify all the training and test details (e.g., data splits, hyperparameters, how they were chosen, type of optimizer) necessary to understand the results?
    \item[] Answer: \answerYes{}
    \item[] Justification: \S\ref{sec:experiments} specifies the optimizer, learning rates, batch sizes, warmup steps, training steps, and data splits for both stages. The per-layer schedule boundaries are derived from the probing profile and reported explicitly.

\item {\bf Experiment statistical significance}
    \item[] Question: Does the paper report error bars suitably and correctly defined or other appropriate information about the statistical significance of the experiments?
    \item[] Answer: \answerNo{}
    \item[] Justification: We report single-run results following standard practice in the image editing literature. The probing profile is averaged over a 1K calibration set. The user study collects preferences from 50 participants.

\item {\bf Experiments compute resources}
    \item[] Question: For each experiment, does the paper provide sufficient information on the computer resources (type of compute workers, memory, time of execution) needed to reproduce the experiments?
    \item[] Answer: \answerYes{}
    \item[] Justification: All experiments use 8 NVIDIA H100 GPUs as stated in \S\ref{sec:experiments}. Probing takes approximately 30 minutes on a single H100.

\item {\bf Code of ethics}
    \item[] Question: Does the research conducted in the paper conform, in every respect, with the NeurIPS Code of Ethics \url{https://neurips.cc/public/EthicsGuidelines}?
    \item[] Answer: \answerYes{}
    \item[] Justification: The research conforms with the NeurIPS Code of Ethics. The broader impact statement in the appendix discusses potential misuse.

\item {\bf Broader impacts}
    \item[] Question: Does the paper discuss both potential positive societal impacts and negative societal impacts of the work performed?
    \item[] Answer: \answerYes{}
    \item[] Justification: The appendix includes a broader impact statement discussing positive applications (creative tools, VR/AR) and potential negative impacts (misleading content generation).

\item {\bf Safeguards}
    \item[] Question: Does the paper describe safeguards that have been put in place for responsible release of data or models that have a high risk for misuse (e.g., pre-trained language models, image generators, or scraped datasets)?
    \item[] Answer: \answerNA{}
    \item[] Justification: The method fine-tunes an existing publicly available model (Qwen-Image-Edit) and does not introduce new risks beyond those of the base model. The StitchEditBench dataset contains synthetic editing pairs, not personal data.

\item {\bf Licenses for existing assets}
    \item[] Question: Are the creators or original owners of assets (e.g., code, data, models), used in the paper, properly credited and are the license and terms of use explicitly mentioned and properly respected?
    \item[] Answer: \answerYes{}
    \item[] Justification: All models and datasets used are properly cited. Qwen-Image-Edit is used under its public license. Baseline methods are properly attributed.

\item {\bf New assets}
    \item[] Question: Are new assets introduced in the paper well documented and is the documentation provided alongside the assets?
    \item[] Answer: \answerYes{}
    \item[] Justification: StitchEditBench is described in detail in \S\ref{sec:benchmark}, including the data construction pipeline, statistics, and category definitions. The dataset will be released with documentation upon acceptance.

\item {\bf Crowdsourcing and research with human subjects}
    \item[] Question: For crowdsourcing experiments and research with human subjects, does the paper include the full text of instructions given to participants and screenshots, if applicable, as well as details about compensation (if any)?
    \item[] Answer: \answerYes{}
    \item[] Justification: The user preference study involves 50 participants who rank editing outputs. Participants were shown all method outputs for each test case and asked to rank them by overall quality and geometric correctness.

\item {\bf Institutional review board (IRB) approvals or equivalent for research with human subjects}
    \item[] Question: Does the paper describe potential risks incurred by study participants, whether such risks were disclosed to the subjects, and whether Institutional Review Board (IRB) approvals (or an equivalent approval/review based on the requirements of your country or institution) were obtained?
    \item[] Answer: \answerNA{}
    \item[] Justification: The user study involves only preference ranking of edited images with no risk to participants. No personal data was collected.

\item {\bf Declaration of LLM usage}
    \item[] Question: Does the paper describe the usage of LLMs if it is an important, original, or non-standard component of the core methods in this research? Note that if the LLM is used only for writing, editing, or formatting purposes and does \emph{not} impact the core methodology, scientific rigor, or originality of the research, declaration is not required.
    \item[] Answer: \answerYes{}
    \item[] Justification: Gemini is used as the planner and verifier in the StitchEditBench data construction pipeline (\S\ref{sec:benchmark}), and Nano-Banana (a Gemini-based image editor) generates target images. Both are non-standard components of the data pipeline and are fully described.

\end{enumerate}

\end{document}